\documentclass[sigconf]{acmart}
\usepackage{enumitem}
\usepackage{algorithm}
\usepackage{algpseudocode}
\usepackage{multirow}
\usepackage{bm}
\usepackage{subcaption}
\AtBeginDocument{%
  \providecommand\BibTeX{{%
    \normalfont B\kern-0.5em{\scshape i\kern-0.25em b}\kern-0.8em\TeX}}}

\copyrightyear{2024}
\acmYear{2024}
\setcopyright{rightsretained}
\acmConference[KDD '24] {Proceedings of the 30th ACM SIGKDD Conference on Knowledge Discovery and Data Mining }{August 25--29, 2024}{Barcelona, Spain.}
\acmBooktitle{Proceedings of the 30th ACM SIGKDD Conference on Knowledge Discovery and Data Mining (KDD '24), August 25--29, 2024, Barcelona, Spain}
\acmISBN{979-8-4007-0490-1/24/08}
\acmDOI{10.1145/3637528.3671959}

\begin{document}

%%
%% The "title" command has an optional parameter,
%% allowing the author to define a "short title" to be used in page headers.
\title{Explanatory Model Monitoring to Understand the Effects of Feature Shifts on Performance}
%%
%% The "author" command and its associated commands are used to define
%% the authors and their affiliations.
%% Of note is the shared affiliation of the first two authors, and the
%% "authornote" and "authornotemark" commands
%% used to denote shared contribution to the research.
\author{Thomas Decker}
\authornote{Both authors contributed equally to this research.}
\orcid{0009-0003-4868-988X}
\affiliation{%
  \institution{Ludwig-Maximilians-Universität}
    \city{Munich}
  \country{Germany} 
}
\affiliation{%
\institution{Siemens AG}
  \city{Munich}
  \country{Germany} 
}
\email{thomas.decker@siemens.com}
\author{Alexander Koebler}
\authornotemark[1]
\orcid{0009-0009-1505-0215}
\affiliation{%
  \institution{Goethe University Frankfurt}
  \city{Frankfurt}
  \country{Germany} 
}
\affiliation{%
  \institution{Siemens AG}
  \city{Munich}
  \country{Germany} 
}
\email{alexander.koebler@siemens.com}
\author{Michael Lebacher}
\orcid{0000-0003-2984-7451}
\affiliation{%
  \institution{Siemens AG}
  \city{Munich}
  \country{Germany} 
}
\email{michael.lebacher@siemens.com}
\author{Ingo Thon}
\orcid{0009-0007-0918-3965}
\affiliation{%
  \institution{Siemens AG}
  \city{Munich}
  \country{Germany} 
}
\email{ingo.thon@siemens.com}
\author{Volker Tresp}
\orcid{0000-0001-9428-3686}
\affiliation{%
  \institution{Ludwig-Maximilians-Universität}
  \city{Munich}
  \country{Germany} 
}
\affiliation{%
  \institution{Munich Center for Machine Learning}
  \city{Munich}
  \country{Germany} 
}
\email{volker.tresp@lmu.de}
\author{Florian Buettner}
\orcid{0000-0001-5587-6761}

\affiliation{%
  \institution{Goethe University Frankfurt}
  \city{Frankfurt}
  \country{Germany} 
}
\affiliation{%
  \institution{Siemens AG}
  \city{Munich}
  \country{Germany} 
}
\additionalaffiliation{%
\institution{German Cancer Research Center}
\city{Heidelberg}
\country{Germany}
 }
% \additionalaffiliation{%
% \institution{German Cancer Consortium (DKTK)}
% \city{Heidelberg}
% \country{Germany}
% }
\email{florian.buettner@dkfz.de}

%%
%% By default, the full list of authors will be used in the page
%% headers. Often, this list is too long, and will overlap
%% other information printed in the page headers. This command allows
%% the author to define a more concise list
%% of authors' names for this purpose.
\renewcommand{\shortauthors}{Thomas Decker et al.}

%%
%% The abstract is a short summary of the work to be presented in the
%% article.
\begin{abstract}
Monitoring and maintaining machine learning models are among the most critical challenges in translating recent advances in the field into real-world applications. However, current monitoring methods lack the capability of provide actionable insights answering the question of why the performance of a particular model really degraded. In this work, we propose a novel approach to explain the behavior of a black-box model under feature shifts by attributing an estimated performance change to interpretable input characteristics. We refer to our method that combines concepts from Optimal Transport and Shapley Values as Explanatory Performance Estimation (XPE). We analyze the underlying assumptions and demonstrate the superiority of our approach over several baselines on different data sets across various data modalities such as images, audio, and tabular data. We also indicate how the generated results can lead to valuable insights, enabling explanatory model monitoring by revealing potential root causes for model deterioration and guiding toward actionable countermeasures.
\end{abstract}
%%
%% The code below is generated by the tool at http://dl.acm.org/ccs.cfm.
%% Please copy and paste the code instead of the example below.
%%
%%
%% Keywords. The author(s) should pick words that accurately describe
%% the work being presented. Separate the keywords with commas.

%% A "teaser" image appears between the author and affiliation
%% information and the body of the document, and typically spans the
%% page.
%\begin{teaserfigure}
%\includegraphics[width=\textwidth]{sampleteaser} 
% \caption{Seattle Mariners at Spring Training, 2010.}
% \Description{Enjoying the baseball game from the third-base
% seats. Ichiro Suzuki preparing to bat.}
%\label{fig:teaser}
%\end{teaserfigure}
%\received{20 February 2007}
%\received[revised]{12 March 2009}
%\received[accepted]{5 June 2009}
%%
%% This command processes the author and affiliation and title
%% information and builds the first part of the formatted document.
\maketitle

\begin{figure}[t!]
    \centering
    \includegraphics[width=\columnwidth]{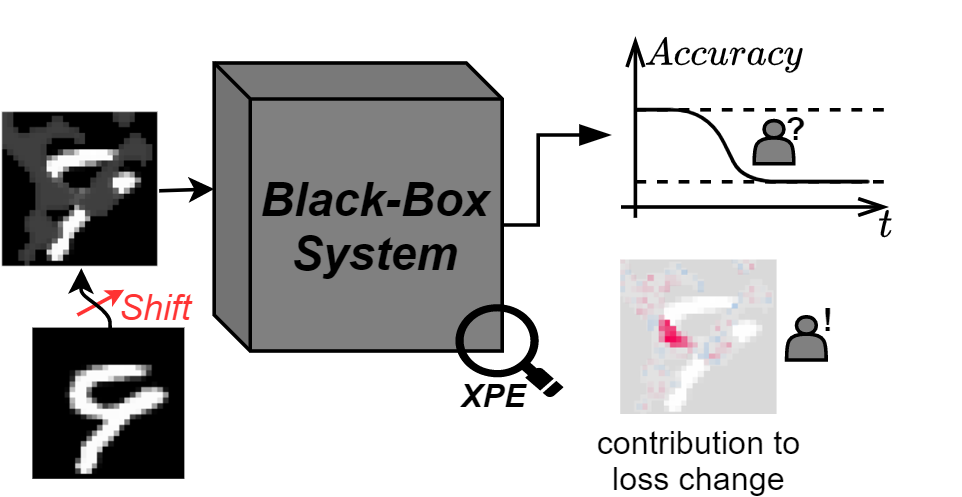}
    \caption{Illustration of an opaque vision system subject to hardware degradation or environment changes, e.g., speckles on the lens or stray light. Explanatory Performance Estimation (XPE) allows to anticipate and explain the resulting performance decrease by highlighting which parts of the shift are harmful. This provides actionable insights to restore performance and facilitate effective model maintenance.}
    \label{fig:fig1}
\end{figure}
\section{Introduction}
Deploying Machine Learning (ML) models successfully in practice is a challenging endeavor as it requires models to cope well with complex and dynamic real-world environments \cite{sculley2015hidden, paleyes2022challenges}. As a consequence, monitoring and maintaining ML-models has been established as a central pillar of the modern ML-Life cycle \cite{studer2021towards, makinen2021needs} and commercial ML frameworks \cite{sagemaker_monitor}. A crucial assumption to assure the validity of a model is that the data distribution during training matches the real-time distribution during deployment. However, this assumption might be violated in real-world applications for various reasons. For instance, data integrity issues such as hardware deterioration or modifications in the collection and processing pipeline could cause a mismatch as well as intrinsic changes in the data generation process due to novel real-world circumstances. Since any potential discrepancy might compromise the reliability of predictions, continuous assessment of the model and its input data is required. For this purpose, many different approaches have been proposed \cite{quinonero2008dataset} that can conceptually be divided into two main categories. Performance monitoring methods \cite{gama2014survey, lu2018learning} enable systematic tracking of the model performance over time and provide an early indication of significant model deterioration. However, such approaches typically require access to ground truth labels at inference time, which is usually infeasible or rather expensive to obtain. In contrast, unsupervised data drift detection \cite{rabanser2019failing, gemaque2020overview} quantifies to which extent the input data characteristics have changed to identify distribution shifts irrespective of the actual performance. While such approaches can help reveal potential problems and prevent unreliable predictions, they are not capable of providing truly actionable insight about why the model might be invalid, how a distribution shift specifically harms the model, and how to optimally counteract it \cite{shergadwala2022human, decker2023thousand}.
More specifically, many different root causes could underlie an observed distribution shift with individual implications. While expensive retraining might be unavoidable in case of an intrinsic change in the relationship between input data and output labels, it would be ineffective for mitigating problems arising from hardware failure such as a defective sensor. Moreover, due to modern machine learning systems' complex and opaque nature, it remains unclear which characteristics of the shifted distribution are responsible for the model failure and we refer to this task as quantifying feature shift importance. In this work, we introduce a new approach coined Explanatory Performance Estimation (XPE) that systematically addresses the desired needs of actionable model monitoring in practice. In particular, we propose a framework that anticipates the performance change caused by an observed distribution shift and guides experts toward potential root causes and actions. More specifically, we make the following contributions: 
\begin{itemize}[leftmargin=18pt]
    \item  We propose Explanatory Performance Estimation (XPE) as a novel and agnostic approach to reveal the specific features through which an observed distribution shift affects model performance in the absence of labels.
    \item  We define multiple innovative metrics to quantitatively evaluate the efficacy of feature shift importance methods in improving model monitoring.
    \item We demonstrate that XPE outperforms several baselines via extensive experiments mimicking real-world shifts such as data quality issues, hardware degradation, and selection bias on three different data modalities. 
    
\end{itemize}
\section{Problem Setting}
We consider the common situation where a machine learning model $f:\mathcal{X}\rightarrow \mathcal{Y}$ has been trained to perform a prediction task in a supervised fashion based on labeled training data $\{(x_s^i,y_s^i)\}_{i=1}^{n_s}$. Further, we assume that the training data originates from a source distribution denoted as $P_s(X,Y)$. At some point during deployment, we suppose that the underlying data distribution changes and further equals to the target distribution $P_t(X,Y)$ with $P_t(X,Y) \neq P_s(X,Y)$. As common in practice, we suppose that during deployment we only have access to unlabeled data instances $\{x_t^i\}_{i=1}^{n_t}$ originating from the marginal target distribution $P_t(X)$. 
As a motivational example, consider the situation in Figure \ref{fig:fig1} where a black-box model processing images is monitored. During deployment, the data distribution changes as a consequence of hardware degradation, e.g., speckles on the camera lens, causing a feature drift in some image areas.
However, the degree of model robustness under a distribution shift might vary across image areas and individual features might be more important to the system in general. Hence, a shift might actually hurt the model's performance only through very specific image areas and revealing those provides valuable information for efficient model monitoring and maintenance. \\
To achieve this, XPE must accomplish two things in a data and model-agnostic manner. First, it should estimate the model's performance under $P_t(X,Y)$ despite missing target labels. 
Second, it should attribute the anticipated performance change to specific input features through which the distribution shift affects the model.
\section{Background and Related Work} 
\paragraph{Optimal Transport}
Optimal transport \cite{villani2009optimal, peyre2019computational} refers to the mathematical problem of identifying the most cost-efficient way to transform one probability measure $\mu$ into another one $\nu$. Consider two measurable spaces $(\mathcal{X}_1, \mu)$ and $(\mathcal{X}_2, \nu)$ and a non-negative cost function $c: \mathcal{X}_1\times \mathcal{X}_2 \rightarrow \mathbb{R}^+$. The Monge formulation aims to find a deterministic transportation map $T: \mathcal{X}_1\rightarrow \mathcal{X}_2$ solving
\begin{align*}
	\inf_{T} \int _{\mathcal{X}_1} c(x, T(x)) d\mu(x) \quad \text{s.t.} \quad T_{\#}\mu = \nu
\end{align*} where $T_{\#}\mu$ describes the push forward measure resulting from probability mass transfer from $\mu$ with respect to $T$, so $T_{\#}\mu(x)=\mu(T^{-1}(x))$. In general, the existence of such a deterministic map is not guaranteed, but the relationship between $\mu$ and $\nu$ can also be expressed via a probabilistic coupling. Mathematically, such a coupling $\pi$ represents any joint distributions over $( \mathcal{X}_1\times \mathcal{X}_2)$ with marginals equal to $\mu$ and $\nu$. This leads to a relaxed problem corresponding to the Kantorovich formulation of optimal transport:
\begin{align*}
 \inf_{\pi} \int_{\mathcal{X}_1\times \mathcal{X}_2} c(x_1,x_2) d\pi(x_1,x_2)
\end{align*}
Note, that a cost-optimal coupling can be shown to exist under mild theoretical assumptions \cite{villani2009optimal} and can easily be estimated based on empirical samples via linear programming \cite{peyre2019computational}. More advanced estimation techniques have been proposed to ensure scalability \cite{NIPS2013_af21d0c9, genevay2016stochastic, lin2019efficient}, to account for known structure within the data  \cite{alvarez2018structured, titouan2019optimal,alvarez2019towards} or to make the estimation more robust  \cite{paty2019subspace, petrovich2020feature, balaji2020robust}. 
Concepts and tools originating from optimal transport theory have been applied to various areas of machine learning concerned with modeling and relating data distributions. This covers for instance generative models \cite{arjovsky2017wasserstein,salimans2018improving, khrulkov2022understanding, rout2022generative} or the evaluation of the semantic correspondence between documents \cite{kusner2015word, huang2016supervised, yurochkin2019hierarchical}, images \cite{liu2020semantic, zhao2021towards, korotin2023neural} and spatio-temporal data \cite{janati2020spatio}.
Another related application of optimal transport is unsupervised domain adaptation \cite{courty2016optimal,courty2017joint,damodaran2018deepjdot}, where the goal is to leverage labeled data from a source domain to obtain models that generalize well to an unlabeled target domain. There also exist proper extensions for cases where a limited number of target labels is available \cite{courty2014domain, yan2018semi} or to improve adaptation results in the presence of general types of data shifts \cite{rakotomamonjy2022optimal,kirchmeyer2022mapping}.
The utilization of optimal transport that is most related to our work is given by \cite{kulinski2023towards}. The authors propose to estimate interpretable transport plans to rather explain the nature of distribution shifts irrespective of their actual effect on a given model.
\paragraph{Feature attribution methods}
Feature attribution methods quantify to what extent individual input features have contributed to a model output of interest and have been established as a popular mean to make the predictions of black-box models more transparent \cite{adadi2018peeking}. More specifically, such methods aim to determine an importance vector $\phi$ such that $\phi_i$ quantifies the influence that each input feature $x_i$ had on a scalar-valued model output. Removal-based methods \cite{covert2021explaining} are a particular kind of attribution techniques that achieve this by simulating the absence of features and evaluating the resulting prediction changes using different computational approaches. A prominent example are Shapley Values \cite{lundberg2017unified}, which have been introduced as a fair way to distribute the total outcome of a coalition game to individual players $D=\{1,\dots d\}$. In this context, a game can be specified via a value function $v(K): 2^D \rightarrow \mathbb{R}$ that quantifies the value that each possible subset or coalition of players $K\subseteq D$ would achieve if only they would contribute. Given a value function, the Shapley Value of each player $i \in D$ results as a weighted average of its marginal contributions over all possible coalitions and orders:
\begin{align*}
 \phi_i = \sum\limits_{K\subseteq D\setminus \{i\}}\dfrac{1}{\binom{d-1}{|K|}d} \left(v(K \cup \{i\}) - v(K) \right)
\end{align*}  
To attain feature attribution for a model $f$ and input $x\in \mathbb{R}^d$, individual features $x_i,\dots, x_d$ resemble players and $v(K)$ defines the hypothetical model prediction where only features in $K\subset [d]$ would be present. Different computational methods have been proposed to enable such a value function by simulating model predictions under partial feature absence \cite{sundararajan2020many, covert2021explaining}.
A prominent one is to integrate affected features out based on the marginal data distribution. For an index set $K$ and decomposition $x = (x_K, x_{K^c})$, this results in
\begin{align*}
    v_{P}(K) &= \int f(x_K, X_{K^c})d P(X_{K^c}) \approx \sum_{i=1}^m f(x_K, x_{K^c}^{(i)} )
\end{align*}
which can easily be estimated using $\{x_i\}_{i=1}^m$ samples from $P(X)$.\\
A popular alternative is to simply replace features with a single uninformative baseline value $\bar{x}$ such that:
\begin{align*}
        v_{\bar{x}}(K) =  f(x_K, \bar{x}_{K^c}).
\end{align*}
Based on such strategies for feature removal, different algorithmic approaches have been developed to compute corresponding feature attributions efficiently \cite{chen2023algorithms}. Moreover, Shapley Values have been shown to satisfy desirable theoretical properties related to fair credit allocations in game theory and are model and data-agnostic. Hence, they can automatically be applied to explain any machine learning model and any kind of model output. In particular, they can also directly be used to explain how individual features have contributed to the overall model performance \cite{covert2020understanding, lundberg2020local}. As a consequence, Shapley Values enjoy great popularity among practitioners \cite{bhatt2020explainable} and they have further been employed for a wide variety of different applications related to machine learning \cite{rozemberczki2022shapley}. 
\paragraph{Feature attributions for model monitoring}
Major ML frameworks such as Amazon SageMaker Model Monitor \cite{sagemaker_monitor} or Google Vertex AI Model Monitoring \cite{taly2021monitoring} offer monitoring of feature attributions and interpret changing importance scores as an indicator for potential performance degradation.  \citet{mougan2022explanation} demonstrate on synthetic tabular examples that monitoring attribution results can be superior compared to monitoring input data characteristics. However, it remains unclear under which circumstances this approach can reliably signal an actual performance decrease and it does not provide any insights regarding why the model might have deteriorated. Another approach is to simply combine drift detection with attribution methods and expect a performance change if an important feature shifts \cite{kenthapadi2022model}. But this can be misleading when the model is robust to the shift occurred in these features. On top of that, attribution methods themselves can be sensitive to shifts and small perturbation \cite{kindermans2019reliability, ghorbani2019interpretation, slack2020fooling}, so simply evaluating them on drifted data may produce unreliable results.
\citet{pmlr-v130-budhathoki21a} leverage Shapley Values to identify potential reasons for a distribution shift based on causal graphs and \cite{pmlr-v202-zhang23ai} apply this idea in the context of model monitoring. While theoretically appealing, these approaches heavily rely on complete knowledge about the causal mechanisms of the true data-generating process which is infeasible to attain in practice. Therefore, these methods cannot be utilized to support monitoring in general deployment scenarios. 
\paragraph{Label-free performance estimation} Reliably estimating the performance of a machine learning model under distribution shifts in the absence of target labels is a significant yet considerably hard problem. In fact, the task has been shown to be impossible when allowing for arbitrary shifts \cite{david2010impossibility, ben2012hardness, garg2022ATC}. However, it becomes feasible if additional assumptions can be posed either on the scope of the shift, the relationship between source and target domain or on the model itself. For instance, importance weighting can be used to estimate the target performance if the experienced change is either due to a covariate shift \cite{sugiyama2008direct, shimodaira2000improving, chen2021mandoline} or resembles a label shift \cite{lipton2018detecting, garg2020unified}. In cases where the model outputs probabilistic predictions that are well calibrated \cite{guo2017calibration, gruber2022better} on the target domain, the corresponding performance can be anticipated purely based on the model’s confidence \cite{guillory2021predicting, jiang2022assessing}. But calibration under domain shifts is particularly challenging and hence an area of active research \cite{ovadia2019can, tomani2021towards, tomani2021post}.
\section{Quantifying Feature Shift Importance}
In this section, we formalize the task of feature shift importance, which aims at revealing through which specific features an observed distribution shift affects a model.
\begin{figure*}[htb!]
    \centering
    \includegraphics[width=\textwidth]{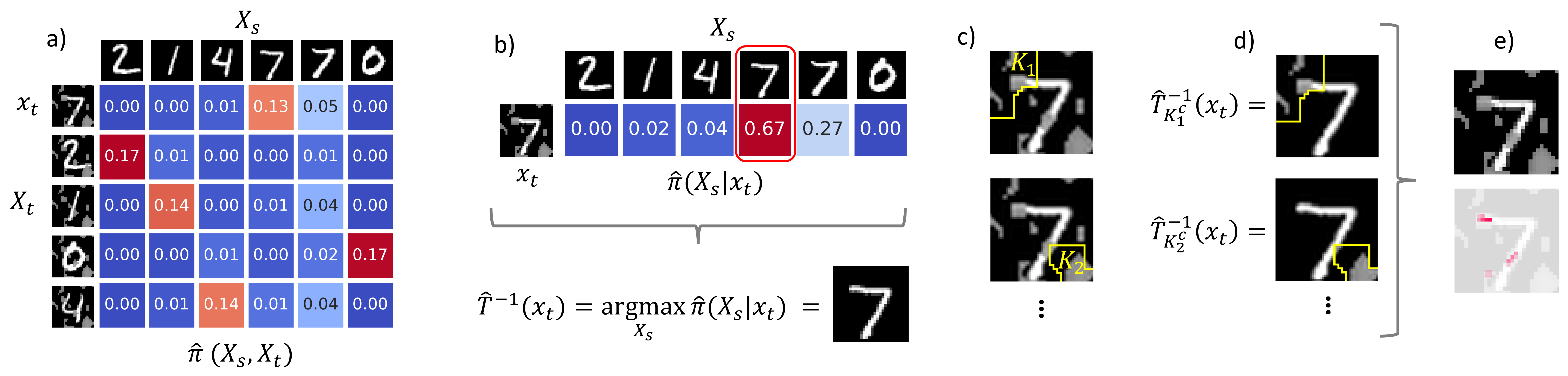}
    \caption{Overview of our proposed framework for Explanatory Performance Estimation (XPE): \textbf{a)} Based on Optimal Transport, an optimal coupling is estimated to sample-wise align empirical source and target distributions. \textbf{b)} For a given target sample $x_t$ the conditional coupling $\hat{\pi}(X_s|x_t)$ indicates the most likely version of $x_t$ in the source domain denoted by $\hat{T}^{-1}(x_t)$. \textbf{c)} Given pre and post-shift version $\hat{T}^{-1}(x_t)$ and $x_t$ one can restrict shifts to individual input feature subsets $K_i$ and \textbf{d)} simulate partial feature shifts $\hat{T}_{K_i^c}^{-1}(x_t)$ by replacing $x_t$ with $\hat{T}^{-1}(x_t)$ outside the considered regions. \textbf{e)} Finally, all simulated partial feature shifts can be aggregated to quantify how individual feature shifts have contributed to the anticipated model loss based on Shapley Values.}
    \label{fig:fig2}
\end{figure*}
\paragraph{Aligning distributions via optimal transport} 
A natural way to gain a better understanding of how a distribution shift precisely impacts the predictions of a machine learning model is to systematically compare its individual predictions before and after the shift happened. For this purpose, suppose that the experienced distribution shift can be expressed by a functional transformation $T$. More precisely, this means that the target domain distribution $P_t$ can be obtained from the source distribution $P_s$ via push-forward operation: $P_t = T_{\#}P_s$.  In this case, the immediate effect on a single model prediction $f(x)$ that is purely induced by the distribution shift can be analyzed by comparing the corresponding predictions $f(x)$ and $f(T(x))$. Modern deep neural networks have demonstrated impressive capabilities to parameterize functional transformations that perform complex and realistic distribution shifts \cite{papamakarios2021normalizing,pang2021image}. However, they typically require a lot of data to be trained and might even rely on labeled source/target pairs. During deployment, there is usually only a limited number of discrete samples from the source and target domain available. Therefore, optimal transport provides a more suitable way to estimate the relationships between $P_s$ and $P_t$ in practice. Suppose we have $n_s$ samples randomly drawn from the source domain $\Omega_s = \{x_s^i\}_{i=1}^{n_s}$ and $n_t$ from the target domain $\Omega_t= \{x_t^i\}_{i=1}^{n_t}$.
Let $\delta_x$ be the Dirac delta function, describing a valid probability distribution concentrated at the point $x$, then the empirical source and target distributions are given by:
\begin{align*}
    &\hat{p}_s = \sum_{x_s \in \Omega_s} \frac{1}{n_s} \delta_{x_s}
    &\hat{p}_t  = \sum_{x_t \in \Omega_t} \frac{1}{n_t} \delta_{x_t}
\end{align*}
The space of possible couplings comprises $\Pi =\{\pi \in \mathbb{R}^{n_s \times n_t}\; |\; \pi \mathbf{1}_{n_t} = \hat{p}_s,  \pi^T\mathbf{1}_{n_s} = \hat{p}_t\}$ and the discrete Kantorovich formulation to match source and target samples via optimal transport yields an optimal coupling $\hat{\pi}$:
\begin{align*}
	\hat{\pi} = \arg \min_{\pi \in \Pi} \sum_{x_s \in \Omega_s }\sum_{x_t \in \Omega_t } c(x_s, x_t) \pi(x_s,x_t)
\end{align*}
Intuitively, $\hat{\pi}$ provides a probabilistic estimate of how samples of the source domain are likely to look in the target domain and vice versa (see Fig. 2a). This equips us with an appealing tool to comprehend the precise nature of the shift and can further be utilized to reveal how an observed shift affected a model. In this case, understanding the impact of a distribution shift for a single prediction $f(x_t)$ could be achieved by comparing it with all predictions corresponding to the potential source version of $x_t$ as implied by the conditional coupling $\hat{\pi}(X_s|x_t)$ (Fig.\ref{fig:fig2}b). Moreover, it is straightforward to transform a probabilistic alignment into a deterministic one by matching each source sample with its most related target sample. This results in a transform $\hat{T}(x_s)=\arg\max_{x_t \in \Omega_t} \hat{\pi}(x_t|x_s)$ and equivalently $\hat{T}^{-1}(x_t)=\arg\max_{x_s \in \Omega_s} \hat{\pi}(x_s|x_t)$ mapping each $x_t$ onto its most likely source version.
In general, the resulting coupling depends on the chosen cost function $c$, but the squared Euclidean distance is a popular default choice for various applications including domain adaptation \cite{courty2016optimal}. In contrast to domain adaptation, we rather seek to understand how a shift impacts a model in order to determine if and specifically why adaptation during deployment might be necessary. To achieve this, we propose a novel way of combining the sample-wise alignment of $P_t$ and $P_s$ implied by optimal couplings with feature attribution methods. 
\paragraph{Shapley Values for Feature Shift Importance}
In order to better understand how an observed input feature shift influenced a model prediction we propose to consider a novel coalition game where the value function $v(K)$ expresses the model prediction under the assumption that only features in $K$ experienced the shift (Fig.\ref{fig:fig2}c). As introduced above, Optimal Transport allows us to identify potential pre- and post-shift versions of data instances related to the empirical source and target distributions. The corresponding results can directly be utilized to perform the required partial distribution shifts of a given target sample $x_t$ (Fig.\ref{fig:fig2}d). Leveraging the probabilistic relationship expressed by the transport coupling $\pi$ we get:
\begin{align*}
    v_{\pi}(K)&=\int f\big(x_{t,K}, X_{s, K^c} \big)d\pi(X_{s, K^c}|X_{t,K} = x_{t,K})
\end{align*}
where ${K^c}$ is the complement of the index set $K$ and $x_{t, K}$ denotes all entries of $x_t$ with index in $K$. If the shift is due to a transformation $T$, then the desired value function is given by:
\begin{align*}
    v_{T}(K)&=f(T^{-1}_{K^c}(x_t)) \;\; \text{with}\;\; T^{-1}_{K^c}(x_t) = (x_{t, K}, T^{-1}(x_t)_{K^c})
\end{align*}
If $T$ is directly inferred from a probabilistic coupling $\pi$ like described above, then $v_T$ resembles a computationally more efficient approximation of $v_{\pi}$, which is equivalent if the conditional coupling $\hat{\pi}(x_s|x_t)$ allocates all probability mass to a single source sample. \\
When computing Shapley Values for such value functions, $\phi_i$ can be interpreted as a measure of how the empirical shift in feature $i$ contributed to the shift-related prediction change (Fig.\ref{fig:fig2}e).
Moreover, carefully comparing $v_{\pi}$ and $v_T$ with existing value functions described in Section 3 reveals a close relationship. While $v_P(K)$ and $v_{\bar{x}}(K)$ are designed to compute partial feature absence for basic feature attribution, our proposed versions $v_{\pi}(K)$ and $v_{T}(K)$ are explicitly tailored to obtain feature shift importance by simulating partial feature shifts. This observation ensures that the resulting Shapley Values $\{\phi_i\}_{i=1}^d$ inherit a variety of desirable theoretical properties \cite{sundararajan2020many, janzing2020feature, pmlr-v202-zhang23ai}, efficient computation strategies \cite{chen2023algorithms} and the relationship to other removal-based explanations \cite{covert2021explaining}.
\section{Explanatory Performance Estimation (XPE)}
\paragraph{Anticipating performance changes during deployment}
When a shift occurs, it is critical to reevaluate whether the model still performs sufficiently well under the new circumstances. Reliably computing the performance of a model would require access to corresponding ground truth labels. Such information is typically not available during deployment and usually requires cumbersome manual efforts. However, empirically aligning labeled source samples $\Omega_s$ and unlabeled target samples $\Omega_t$ via transformation $T$ also equips us with a reasonable way to anticipate the performance by supposing that all linked instances exhibit the same label. More precisely, one can obtain for any $x_t \in \Omega_t$ a label estimate $\hat{y}_t$ by allocating the known label of the linked source sample $T^{-1}(x_t) \in \Omega_s$. Under appropriate assumptions on the nature of the shift, this heuristic will accurately estimate the performance in the target domain. 
\begin{definition}\label{th:perf_est}
Let $\big(\mathcal{X}_s\times \mathcal{Y}_s, P_s(\mathcal{X}_s, \mathcal{Y}_s)\big)$, $(\mathcal{X}_t\times \mathcal{Y}_t, P_t(\mathcal{X}_t, \mathcal{Y}_t))$ be two probability spaces corresponding to the source and target domains. For measurable sets $A$, let $d_{TV}(\mu,\nu)=\sup_{A}\lvert \mu(A) - \nu(A) \rvert$ be the statistical total variation distance between two probability measures $\mu$ and $\nu$. A distribution shift from $P_s$ to $P_t$ is  $\bm{\varepsilon}$\textbf{-approximate label-preserving} with respect to $T$ if there exists 
\begin{itemize}[leftmargin=18pt]
    \item a functional Transformation $T: \mathcal{X}_s \rightarrow \mathcal{X}_t$ such that  $P_t(X_t)= P_s(T^{-1}(X_t))$ and additionally
    \item $ d_{TV}\big(P_t(Y|T(x_s)),P_s(Y|x_s)\big) \le  \varepsilon \quad \forall x_s \in \mathcal{X}_s$
\end{itemize} 
\end{definition}
If the observed distribution shift is approximately label-preserving, then the estimated performance via label transport is close to the actual one in the target domain:
\begin{theorem}
Let $\mathcal{L}_t$ be the true target performance of a model $f$ expressed via a loss function $\mathcal{L}$ with $\lVert \mathcal{L} \rVert_{\infty} \le 1$. Denote $\widehat{\mathcal{L}_t^T}$ as estimated target performances resulting from label transfer via transformation $T$. If the observed distribution shift is $\varepsilon$-approximate label-preserving, then:
	\begin{align*}
		\lvert \mathcal{L}_t - \widehat{\mathcal{L}_t^T}\rvert \le 2\varepsilon 
	\end{align*}
\end{theorem}
The proof and further details are given in the Appendix. Intuitively, this establishes a continuity result, showing that if the decision boundary for transported samples in the target domain is close to the original one, then the estimation error can also be expected to be small. This assumption might seem restrictive in general, but we want to highlight that label-free performance estimation only becomes feasible in the presence of additional constraints. Relying specifically on this assumption resonates quite well with typical real-world causes for distribution shifts during deployment: Physical changes in the environment, hardware degradation, or other data quality issues can all be considered as shifts that modify input data characteristics without necessarily affecting the label.
\paragraph{Explanatory Performance Estimation using Shapley Values} Our proposed \textbf{Explanatory Performance Estimation} (XPE) approach combines transport-based label estimation with feature shift importance in the following way. Given a loss function $\mathcal{L}: \mathcal{Y} \times \mathcal{Y} \rightarrow \mathbb{R}^+$ we define a new value function $v^{\textit{XPE}}(K) = \mathcal{L}\left(f\big(\hat{T}_{K^c}^{-1}(x_t)\big), \hat{y}_t \right)$ which expresses directly the anticipated performance change under partial feature shifts. The corresponding Shapley Values $\phi^{\textit{XPE}}$ finally indicate through which specific features an observed distribution shift impacts the anticipated performance, providing valuable information regarding potential root causes of model degradation.
For classification models that output an entire probability distribution over possible classes $C$, we further propose a label-estimation-free variation of XPE, which we call \textbf{Explanatory Performance Proxy Estimation} (XPPE). Let $H(f(x))=-\sum_{c\in C} f_c(x) \log(f_c(x))$ be the entropy of the probabilistic model output $f(x)$ capturing the model's degree of uncertainty about its prediction. By considering higher predictive uncertainty as a proxy for potential model degradation we define a novel value function
$v^{\textit{XPPE}}(K)= H(f(\hat{T}_{K^c}^{-1}(x_t))$. It computes the model uncertainty under partial shifts and the resulting Shapley Values $\phi^{\textit{XPPE}}$ signal through which features a shift has affected the model's confidence, which does not require a label estimate. Thus, XPPE can be superior to XPE when the shift does not satisfy Def. 5.1 of being approximately label preserving.
\section{Experiments}
The goal of our experiments is to rigorously analyze feature shift attributions and their capabilities to reliably explain the model behavior under deployment-related distribution shifts. In total, we report the results on nine different datasets covering three different data modalities with associated data shift scenarios. In Section 6.1, we analyze four different vision tasks and demonstrate that our method is the most effective way to understand a performance decrease caused by shifts resembling hardware degradations. In Section 6.2 we show, for a speech classification task, that only our proposed methods can reliably signal an induced selection bias. Finally, in Section 6.3 we evaluate four tabular data sets and demonstrate the capabilities of our method to correctly assign a performance decline due to data quality issues in the form of missing values. More details are documented in the Appendix and code is provided at \texttt{https://github.com/thomdeck/xpe}.
\paragraph{Baselines} To assess the capabilities of XPE we first formalize baselines that are connected to existing model monitoring practices. Remember that XPE aims to evaluate by which specific features an observed distribution shift impacts the model performance. A first baseline for this purpose is to simply check whether predictions in the target domain tend to rely on other features compared to the source domain. Let $\phi(x)$ be the outcome of standard Shapley Values explaining the prediction $f(x)$ for an instance $x$. Given an estimated transportation map $\hat{T}$, we can simply compare the explanations of two matched samples individually and define a \textbf{local attribution difference} (LAD):
\begin{align*}
   \phi^{\textit{LAD}}(x_t, \hat{T})=\lvert\phi(x_t) - \phi\big(\hat{T}^{-1}(x_t)\big)\rvert
\end{align*} If this difference is large for a specific feature, then the distribution change in this feature might be particularly harmful. Note, that this method also relates to the existing practice of monitoring changes in feature attributions \cite{sagemaker_monitor, taly2021monitoring, mougan2022explanation} proposed to detect model degradation during deployment. A second baseline that does not require alignment of source and target samples is first to perform unsupervised drift detection and consider only shifted features which are also important for the model as intermediaries of the shift. For this purpose, we leverage a two-sided Kolmogorov-Smirnov (KS) test to assess whether the distribution of each feature within the source samples is significantly different from the corresponding one of the target samples. Let $M^{\textit{KS}}\in \{0,1\}^d$ be a binary mask indicating which of the $d$ input features have drifted according to the KS-test. 
Then, the \textbf{Attribution $\times$ Shift} (AxS) baseline reads:
\begin{align*}
    \phi^{\text{AxS}}(x_t, M^{\textit{KS}}) = \phi(x_t)\odot M^{\textit{KS}}
\end{align*}
Note, that this baseline captures the idea of considering only features that have drifted and are simultaneously important for the model as potentially harmful gateways of the observed distribution shift mentioned in \cite{kenthapadi2022model}. 
\paragraph{Defining suitable metrics}
Quantitatively evaluating any kind of model explanation is challenging, but a desirable property is faithfulness \cite{bhatt2021evaluating}. It generally tries to asses if perturbing features with high attribution scores also cause coherent prediction changes and a variety of different related metrics have been proposed \cite{ancona2018towards, yeh2019fidelity, bhatt2021evaluating}. To evaluate feature shift attributions, we reformulate the faithfulness criterion in the following way: When features with high shift attributions are shifted back, we expect the model performance to recover equivalently. Suppose access to the true pre-shift version $T^{-1}(x_t)$ of a target sample $x_t$ as well as to the ground truth source and target labels $y_s$ and $y_t$. Then, we can define \textbf{Shift-Faithfulness} (S-Faith) of a feature shift attribution  $\phi^{\textit{Shift}}$ as the correlation between the actual performance change under partial feature shift and the sum of allocated shift importance:
\begin{align*}
\textit{S-Faith}=\underset{{K\in \binom{d}{|K|}}}{\textit{corr}}\left( \sum_{i \in K} \phi^{\textit{Shift}}_i, \mathcal{L}\big(f(x_t),y_t\big) -\mathcal{L}\big(f\big(T^{-1}_{K}(x_t)\big),y_s\big) \right)
\end{align*}
Here we adapted the metric based on the notation from \cite{bhatt2021evaluating}, where the correlation is computed using different feature subsets $K$ with fixed size $|K|$. Another popular metric to measure the quality of feature attributions is RemOve And Retrain (ROAR) \cite{hooker2019benchmark} assessing if the performance actually decreases when important features are removed and models retrained. Consequently, we propose an adapted metric coined \textbf{remove, retrain, and shift (ROAR-S)}, which evaluates whether the performance decrease caused by a shift diminished if features with high shift importance are removed and the model retrained. More precisely, we define the ROAR-S score as the proportion of shift-induced performance decrease that remains when for each instance the top $5\%$ of input features highlighted by $\phi^{\textit{Shift}}$ are removed, and the model is subsequently retrained. If this score is small, the distribution change no longer affects the performance and the shift importance is reliable. More details about this metric are deferred to the Appendix. To quantify the actionability of explanations we consider the \textbf{Complexity} $(\textit{Cpx})$ metric \cite{bhatt2021evaluating}, which is defined as the Shannon entropy of the normalized attribution values: $\textit{Cpx}(\phi^{\textit{Shift}})=H(|\phi^{\textit{Shift}}|/\sum_i |\phi^{\textit{Shift}}_i|)$. This expresses the uncertainty of shift attribution results across all input features and lower values indicate that the method communicates the potential reason for model degradation more concisely.
\subsection{Understanding the effects of hardware degradation}
For the first experiment we consider a variety of popular lightweight image datasets, including MNIST \cite{lecun1998gradient}, images of fashion items \cite{xiao2017fashion} as well as various types of medical images \cite{medmnistv2}, and simulated several distribution shifts mimicking potential camera-related hardware degradation or physical changes in the environment \cite{Mu2019MNISTCAR}. This setup ensures complete knowledge about the true pre- and post-shift pairs which is crucial for evaluating the quality of shift attribution methods via Shift-Faithfulness. For each dataset, we fitted a LeNet model \cite{lecun1998gradient}, evaluated the metrics based on $500$ test samples and used the cross-entropy loss as a performance measure. The average results are reported in Table 1 and indicate that XPE almost consistently outperforms all baselines followed by XPPE as second best. Most of the time, all other baselines are not correlated at all with the true performance change induced by the shift of highlighted features. The corresponding results for Complexity imply that the explanations of XPE also tend to be the most concise given a sufficient degree of faithfulness. For cases where other methods provide significantly less complex results, they typically have almost no correlation with the actual performance decrease. Moreover, the label transport accuracy $(\hat{y}_t = y_t)$ was for all considered shifts $>85\%$. This is in line with the Theorem above as applying a corruption to an image equals a functional transformation that preserves its label and shows that aligning via optimal transport is capable of apprehending the considered transformations. To confirm our findings, we also evaluated ROAR-S and the results demonstrate that the performance change caused by the shift is on average best mitigated when dropping features according to XPE and XPPE. 
\begin{table*}[t!] \centering
     \resizebox{\textwidth}{!}{
         \begin{tabular}{*{2}{c}|*{2}{c}|*{2}{c}|*{2}{c}|*{2}{c}|*{2}{c}|*{2}{c}|*{2}{c}|*{2}{c}||c }
         \toprule
         & & \multicolumn{2}{c}{brightness} 
         & \multicolumn{2}{c}{contrast} 
        & \multicolumn{2}{c}{dotted} 
        & \multicolumn{2}{c}{fog}  
         & \multicolumn{2}{c}{gaussian} 
        & \multicolumn{2}{c}{impulse} 
        & \multicolumn{2}{c}{spatter} 
        & \multicolumn{2}{c}{zigzag}
        & AVG
        \\
       &  & S-Faith$\uparrow$& Cpx$\downarrow$ & S-Faith$\uparrow$& Cpx$\downarrow$ & S-Faith$\uparrow$& Cpx$\downarrow$ & S-Faith$\uparrow$& Cpx$\downarrow$ & S-Faith$\uparrow$& Cpx$\downarrow$ & S-Faith$\uparrow$& Cpx$\downarrow$ & S-Faith$\uparrow$& Cpx$\downarrow$ & S-Faith$\uparrow$& Cpx$\downarrow$ & ROAR-S$\downarrow$\\
         \midrule
         \multirow{4}{*}{\rotatebox[origin=c]{90}{MNIST}} & XPE & 0.45 & \textbf{4.03} & \textbf{0.53} & 5.91  & \textbf{0.82}  
         & 2.55  & \textbf{0.52} &  5.88 & \textbf{0.71}  & 3.71  & \textbf{0.82} &  \textbf{2.62} & \textbf{0.69} & \textbf{4.32}  & \textbf{0.78} & 3.38 & \textbf{0.37} \\
         & XPPE  & \textbf{0.49}   &  4.69 & 0.48  &  5.67 & 0.70 &  \textbf{2.53}& 0.49  &  5.96 & 0.61  &  \textbf{3.52} & 0.77  & 2.64  &  0.51  &  4.44 &  0.50 & \textbf{3.33} & 0.78\\
         & LAD & 0.01 & 5.62 & 0.32  & \textbf{5.16} & 0.22 & 4.70 & 0.15  & \textbf{5.69}  & 0.11 & 5.25  & 0.16  &  4.67 & 0.21 &  5.05  & 0.13  & 4.82 & 2.12\\
         & AxS & -0.04  & 5.42  & 0.24 & 5.81  & 0.31  & 3.12  & 0.11 & 5.92  &  0.07  & 4.92 & 0.11&  2.73 & 0.26 & 4.91   & 0.21 &  3.82 & 1.87\\
         \midrule
         \multirow{4}{*}{\rotatebox[origin=c]{90}{FashionM}} & XPE & \textbf{0.58} & \textbf{5.86} & \textbf{0.58} & \textbf{5.91}  & \textbf{0.79}  
         & \textbf{2.82}  & \textbf{0.72} &  \textbf{5.97} & \textbf{0.73}  & \textbf{5.24}  & \textbf{0.79} &  3.19 & \textbf{0.72} & \textbf{4.85}  & \textbf{0.82} & \textbf{3.31} &  \textbf{0.20}\\
         & XPPE  & 0.20  & 5.88  & 0.04  &  6.01  & 0.57 &  2.90  & 0.06  & \textbf{5.97} & 0.50  &  5.35 & 0.57  & 3.33   &  0.42  &  4.92  &  0.49 &  3.65 & 0.55\\
         & LAD & 0.00  & 6.12  & 0.06  & 6.05  & 0.04 & 5.63  & -0.00  & 6.11   &  0.04 &  5.96  & 0.02  & 5.57  & 0.03 &  5.82   & 0.05  &  5.67 & 0.77\\
         & AxS & -0.01  &  5.99  & 0.01 & 5.93  & 0.07  & 3.24   & 0.01 & 6.04   &  0.03  & 5.72  & 0.04 & \textbf{2.85}  & 0.09 &  5.04   & 0.07 &  3.88 & 0.78\\
         \midrule
         \multirow{4}{*}{\rotatebox[origin=c]{90}{OrganaM}} & XPE & \textbf{0.34}&5.59&	\textbf{0.33} & 5.44 & \textbf{0.74} & 3.06 &	\textbf{0.27} & 5.61 & \textbf{0.59} & 5.34&	\textbf{0.81} &2.94&	\textbf{0.58} &4.27&	\textbf{0.68} &3.78 & \textbf{0.48}\\
         & XPPE  & 0.10 &5.75& 0.01 & 5.70& 0.48& 3.08&	0.03 &5.71&	0.33 & 5.40& 0.44 &2.89&	0.34 & 4.28& 0.30 & 3.78 &0.64\\
         & LAD & 0.01 &5.63&	0.12 &5.56&	0.03 & 3.08& 0.08 & 5.68& 0.02 &5.70&	0.05 &5.70&	0.03 & 5.68& 0.02 &5.64 &0.60\\
         & AxS & -0.02 &\textbf{4.84}& 0.06 & \textbf{4.80}& 0.09 & \textbf{2.43}& 0.02 &\textbf{5.11}&	0.02& \textbf{5.04}&	0.10 &\textbf{2.57}&	0.04 & \textbf{3.40}& 0.05 &\textbf{3.02} & 0.56\\
         \midrule
         \multirow{4}{*}{\rotatebox[origin=c]{90}{PneumM}} & XPE & \textbf{0.38} &5.88&	\textbf{0.33} &5.17&	\textbf{0.78} &3.21&	\textbf{0.25} &5.57&	\textbf{0.63} &5.79&	\textbf{0.79} &3.15&	\textbf{0.62} &4.43&	\textbf{0.72} &3.90 & 0.96 \\
         & XPPE  & 0.20 &5.87 & 0.17 &5.49&	0.49 &3.19&	0.07 &5.51&	0.43 &5.78&	0.40 &3.14&	0.23 &4.40&	0.45 &3.89 & \textbf{0.92}\\
         & LAD & -0.01 &5.50 & 0.03 &5.08&	0.08 &5.84&	0.01 &5.24&	0.02 &5.83&	0.04 &5.82&	0.03&5.90&	0.01&5.89 & 2.00\\
         & AxS &  -0.01 &\textbf{5.15}&	0.07 &\textbf{4.10}&	0.10 &\textbf{3.03}&	-0.00 &\textbf{4.76}&	0.02 &\textbf{5.48}&	0.08 &\textbf{2.98}&	0.05 &\textbf{3.97}& -0.00&\textbf{3.61} &5.05\\
    
         \bottomrule
     \end{tabular}
     }
     \caption{Average S-Faithfulness (S-Faith), Complexity (Cpx), and ROAR-S results of shift attributions methods for a LeNet on different image datasets and corruptions. A higher S-Faithfulness value indicates that features highlighted by $\phi^{\textit{Shift}}$ are strongly correlated with the true performance change caused by the shift in these features. A lower Complexity value corresponds to more concise explanations and a lower ROAR-S score signals that removing features based on $\phi^{\textit{Shift}}$ effectively mitigates the shift-induced performance change.} \label{tab:1}
     \end{table*}

\begin{figure}[t!]
    \centering
    \includegraphics[width=\columnwidth]{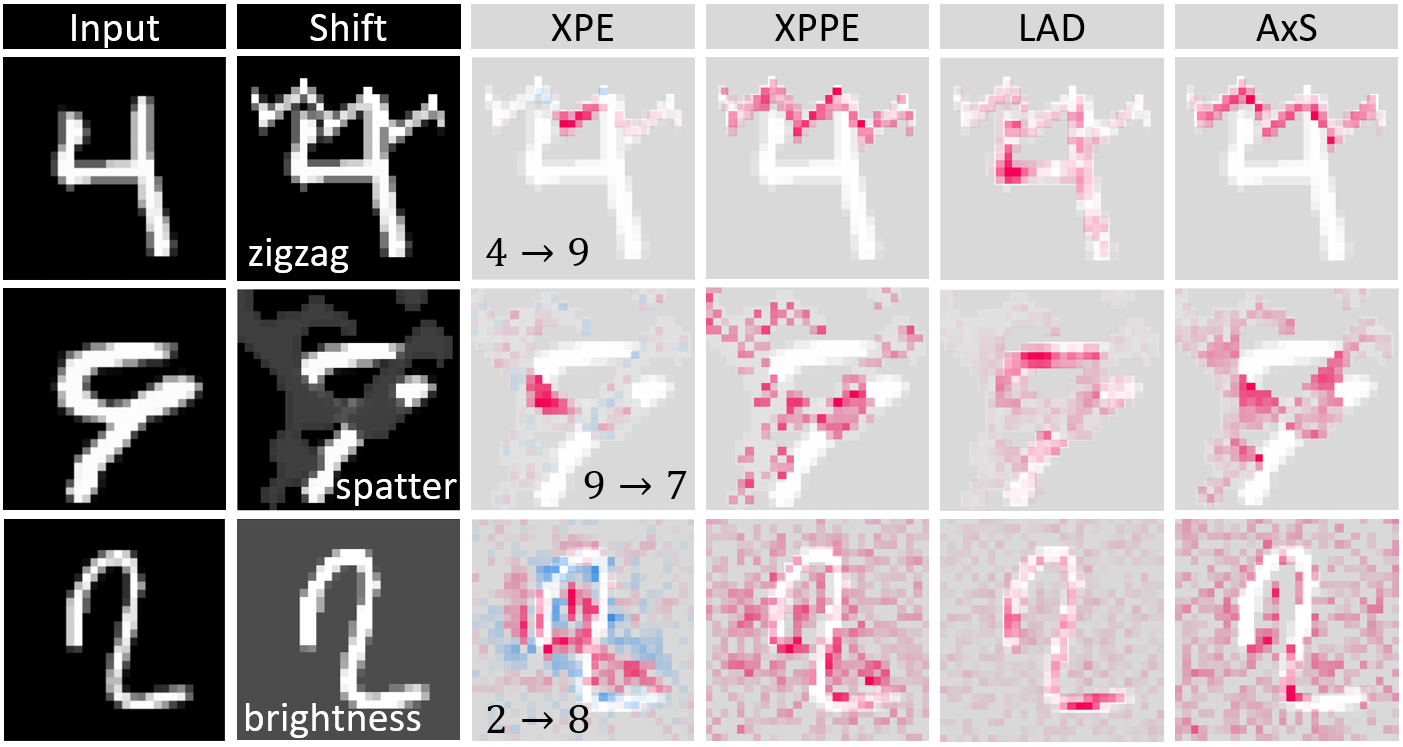}
    \caption{Feature shift importance on MNIST for local and global corruptions. The explanations given by XPE are most intuitive, only highlighting the area of the zigzag connecting the top parts of the '4', which changes the model's prediction to a '9'. A similar observation can be made for the spatter corruption, changing the prediction from '9' $\rightarrow$ '7'. XPE also allows uncovering the image regions shifting the prediction from '2' $\rightarrow$ '8' given a global increase in brightness.}
    \label{fig:mnist_x}
\end{figure}

\begin{figure}[h!]
    \centering
    \includegraphics[width=\columnwidth]{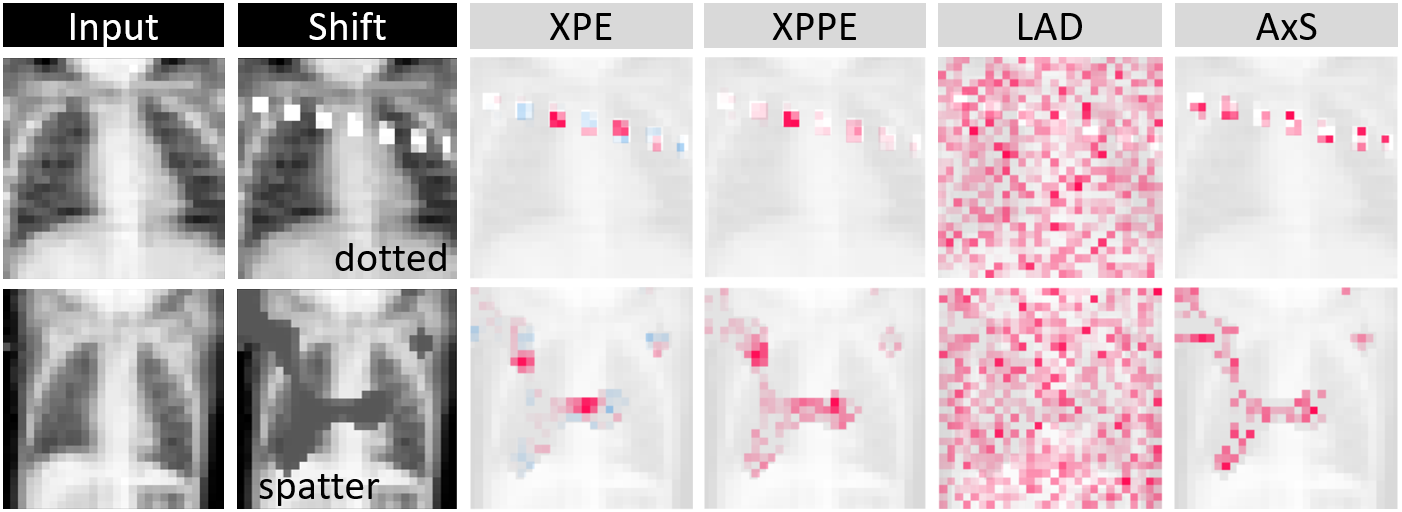}
    \caption{Feature shift importance for pneumonia detection dataset (PneumM) consisting of real chest X-rays. Neither image shows pneumonia. Only XPE consistently assigns the contribution to the loss increase caused by the shift to areas that might resemble pneumonia indicators (white spots) within the lung area.
    }
    \label{fig:pneum_x}
\end{figure}

\begin{figure}[htb!]
    \centering
    \includegraphics[width=\columnwidth]{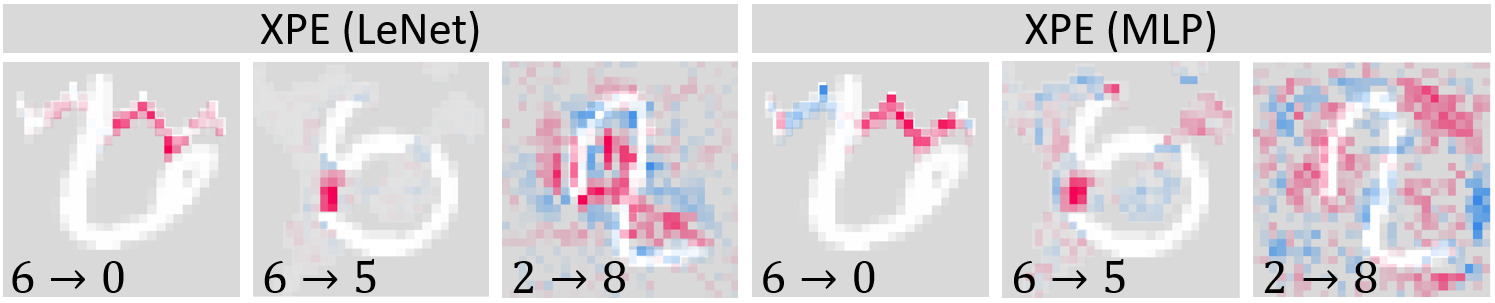}
    \caption{Comparison of XPE for zigzag ('6' $\rightarrow$ '0'), spatter ('6' $\rightarrow$ '5'), and brightness ('2' $\rightarrow$ '8') corruptions between a LeNet and an MLP model.}
    \label{fig:mlplenet}
\end{figure}

\paragraph{Deriving intuitive and actionable insights} Next, we would like to demonstrate how the results obtained via XPE can yield novel and actionable insight into the model behavior under distribution shifts concerning images.
To do so, we locally examine shift attributions on the MNIST digit and the pneumonia detection dataset (PneumM) where a certain shift had a particularly harmful effect on the prediction and seek to understand the reasons. In Figure \ref{fig:mnist_x} and Figure \ref{fig:pneum_x} we plot some of these examples and notice that for such instances, the shifts do indeed perturb essential image regions in a way that suggests a different class, i.e., a different digit or a positive pneumonia classification. By consulting the different shift attribution results to narrow down a concrete reason we see that only XPE consistently highlights the parts of the corruption that actually alter the appearance of a digit towards a different one or resemble white spots in the lung area indicating pneumonia. This shows that also the model is mainly misled by the intuitive regions, which cannot be concluded from the other results. This information can help end-users take efficient and targeted corrective measures for their applications, such as cleaning or repairing the camera lens or removing ambient light sources.
\paragraph{Supporting model selection}
Moreover, such analysis can also be used to compare how different models handle certain distribution shifts. In Figure \ref{fig:mlplenet} we visualize some results of XPE for the LeNet analyzed above and a Multi-Layer-Perceptron (MLP) with one hidden layer. The plots illustrate that the LeNet's predictions mainly get harmed through the shift of features in close proximity to the actual digit, whereas the MLP also gets distracted by parts of the corruptions at irrelevant peripheral regions. This implies that the LeNet model with convolutional feature extraction generally copes better with the considered shifts, while the MLP might has internalized some unstable global patterns. Such results convey valuable insights that can help to disclose model deficiencies, perform model selection, or guide model debugging. 
\subsection{Revealing selection bias in the training set for audio data}
\begin{figure}[t]
    \centering
    \includegraphics[width=\columnwidth]{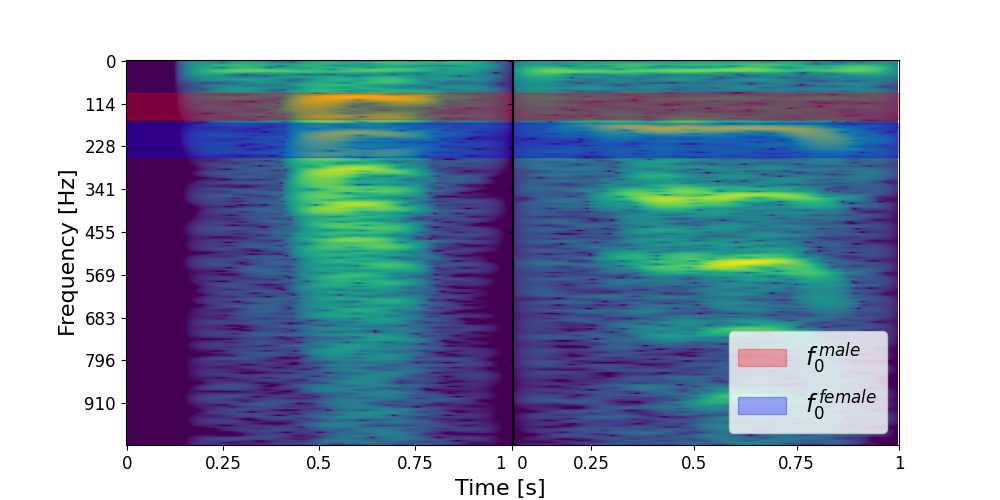}
    \caption{Spectrogram considering frequencies of up to $1024 Hz$ for a male (left) and a female (right) participant pronouncing the digit three. The fundamental frequency bands for males $f_0^{\textit{male}}$ and females $f_0^{\textit{female}}$ are highlighted.
    }
    \label{fig:spec}
\end{figure}

With the application of machine learning models in an increasing number of real-world applications, the identification of biases included in the model becomes more important \cite{10.1145/3457607}. This is particularly relevant in voice recognition and assistant systems, where the performance of a model differs between groups of people formed by gender and region because they are not sufficiently represented in the training data \cite{10.1145/3308560.3317597, chen2022exploring}.\newline
In this experiment, we demonstrate how XPE can help to indicate such a selection bias if relevant meta-information about the included groups in the source and target distribution is absent. We analyze the prediction task of classifying spoken digits '0' to '9' from audio signals based on the dataset provided by \cite{audiomnist2023} containing real voice recordings of 60 different speakers with varying ages, sexes, and accents. Since our method is agnostic to the model used, we follow exactly the same model choice made by the original researchers \cite{audiomnist2023}, which propose to use an AlexNet \cite{AlexNet} as a classifier. To do so, the voice recordings are transformed into 2D spectrograms of size $227\times 227$ using the Short-time Fourier Transform (STFT), and an example is shown in Figure \ref{fig:spec}. To introduce a selection bias, we consider a model solely trained on male participants, so the model faced a significant bias during training with potential consequences on the performance during operation. We evaluated this model on unseen test sets with varying fractions of female participants and asses if feature shift importance can identify the selection bias as a source for model degradation.\\
Due to the high dimensionality of the input data $(d=51.529)$, we apply two general extensions to improve the scalability of all attribution methods and reduce the computational effort. First, we solved the optimal transport problem based on intermediate network activations of the used AlexNet instead of the raw inputs \cite{courty2016optimal}. This allows us to reduce the effective size over which the transport needs to be computed and increases the quality of the resulting performance estimation significantly. Second, we computed shift importance at the level of frequency bands instead of all individual spectrogram values, which also reduces the computational load while maintaining the interpretability of the resulting explanations. To do so, we group features by splitting the frequency axis into $32$ frequency bands of $32Hz$ resolution and aggregating over the entire time axis. This level of granularity is sufficient to distinguish gender-specific characteristics related to the induced selection bias. 
Based on these application-specific improvements, we evaluate all shift importance methods on the biased model (trained on males only) when facing three shift scenarios with varying bias severity. First, a test set consisting of unseen male participants, such that a difference in performance between training and test samples should be predominantly due to a change in non-gender-specific frequencies, e.g., caused by different pronunciations between individual persons. Second, we evaluate the model on a test set with an equal number of male and female participants. Finally, we test the model on a test set consisting only of women, which results in a strong gender-specific shift. To assess the quality of feature shift importance in this context, we build on existing domain knowledge about gender-specific human vocal frequencies \cite{Fitch1970ModalVF}. It suggests that distinctive frequency ranges are between $85Hz$ and $155Hz$ for males and $165Hz$ and $255Hz$ for females. Therefore, we expect that as the fraction of females in the test set increases, the overall shift importance allocated to gender-specific frequencies should also increase proportionally, successfully indicating the severity of the selection bias. To investigate this, we calculate the proportion of feature shift importance within the combined male and female fundamental vocal frequency bands ($\Phi^{\textit{Shift}}_{\textit{gender}}$) relative to the total feature shift importance across all frequencies ($\Phi^{\textit{Shift}}_{\textit{total}}$) as a global measure over the entire test dataset.
\begin{align*}
    \Phi^{\textit{Shift}}_{\textit{ratio}} = \frac{\Phi^{\textit{Shift}}_{\textit{gender}}}{\Phi^{\textit{Shift}}_{\textit{total}}}
\end{align*}
\begin{figure}[t]
    \centering
    \includegraphics[width=\columnwidth]{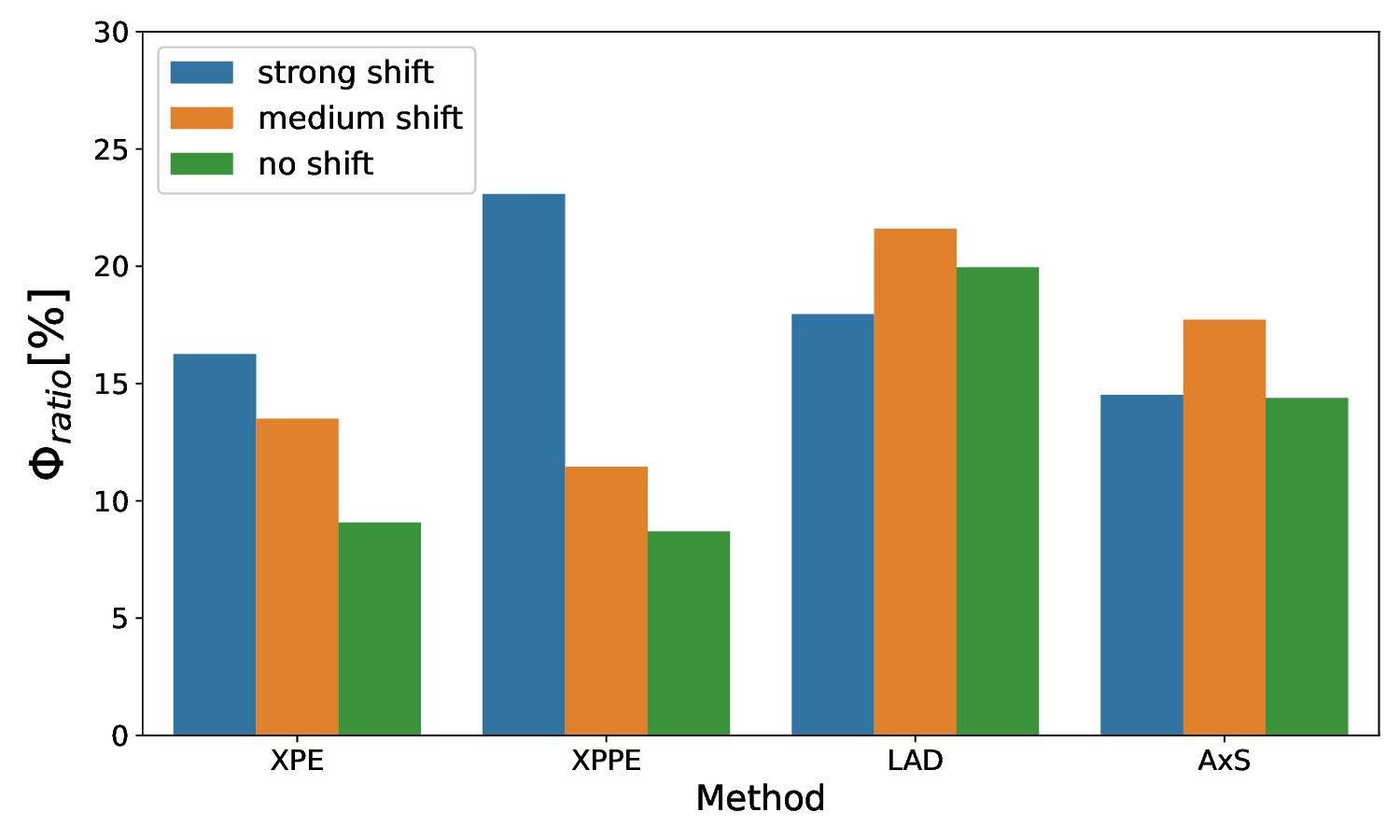}
    \caption{Ratio of feature shift importance allocated to gender-specific frequencies for a biased model solely trained on male participants and evaluated on sets with only women (strong shift), with equal gender distribution (medium shift), and with solely men (no shift) not in the training set. Only XPE and XPPE attribute importance proportionally to the strength of the induced selection bias.}
    \label{fig:audiomnist}
\end{figure}
The results presented in Figure \ref{fig:audiomnist} demonstrate that only XPE and XPPE attribute importance proportionally to the strength of the induced selection bias, while all other baselines are insensitive to the faced distribution shift. This suggests that in cases where a selection bias is introduced during the training of machine learning models, XPE and XPPE can detect it and prompt further investigation, even in the absence of labels in the target domain or corresponding meta-information.

\subsection{Assessing the impact of missing values in tabular data}
As a final experiment, we investigate the effectiveness of shift attribution methods in the presence of data quality issues on four tabular datasets from a popular evaluation benchmark \cite{grinsztajn2022tree}. Specifically, we focus on the commonly encountered problem of facing missing values due to a compromised data acquisition process. This is typically handled via an imputation strategy $\mathcal{I}$ that replaces missing values based on a predefined heuristic such as the feature mean. For each dataset, we fitted two different models (XGBoost \cite{chen2016xgboost} and an MLP) and corrupted a set of $n=500$ unseen test samples $X_t$ by randomly selecting a column $j$ and marking 25\% of values as missing, yielding a perturbed test set $\tilde{X}_t$. We then assess how reliably the shift importance values associated with column $j$, denoted by $\phi^{\textit{Shift}}_j$, correspond to the true impact of the corruption. To this end, we compute the correlation between $\phi^{\textit{Shift}}_j$ and the loss difference due to missing value imputation across all samples in the test set. This results in a global and feature-specific version of Shift-Faithfulness, which we refer to as Global Performance-Correlation (GPC): 
\begin{align*}
    \textit{GPC} (\phi^{\textit{Shift}}_j) =
    \underset{\substack{\tilde{x}_t \in \tilde{X}_t \\ x_t \in X_t}}{\textit{corr}}\left(\phi^{\textit{Shift}}_j(\tilde{x}_t), \mathcal{L}\big(f(\mathcal{I}(\tilde{x}_t)),y_t\big) -\mathcal{L}\big(x_t,y_t\big) \right) 
\end{align*}
Here, $\mathcal{I}(\tilde{x}_t))$ corresponds to the corrupted sample $\tilde{x}_t$, where the missing values are filled using a mean value imputer $\mathcal{I}$ fitted on the training set. While standard S-Faith evaluates the quality of a local shift attribution result for an individual sample, GPC is a global measure of shift attribution fidelity for specific features of interest across an entire test set. The corresponding results in Table 2 demonstrate that XPE and XPPE allocate an importance score to the corrupted column that matches the actual loss contribution caused by the quality issue best. This consolidates the capabilities of our introduced methods to improve model monitoring in the presence of common data quality issues also for tabular data. 

\begin{table}[h] \centering
\resizebox{\columnwidth}{!}{
         \begin{tabular}{c|*{2}{c}|*{2}{c}|*{2}{c}|*{2}{c}}
         \toprule
         &  \multicolumn{2}{c}{covertype} 
         &  \multicolumn{2}{c}{pol}
         & \multicolumn{2}{c}{electricity} 
         & \multicolumn{2}{c}{bank-marketing} 
        \\ 
       & XGB& 
       MLP& 
       XGB&
       MLP&
       XGB&
       MLP& 
       XGB&
       MLP  \\
         \midrule
         XPE &\textbf{0.05} &0.05 &  0.03& \textbf{0.43}  &0.09 & 0.06&\textbf{0.12} & \textbf{0.06} \\
         XPPE &\textbf{0.05} &\textbf{0.13} &  \textbf{0.17} & 0.42 &\textbf{0.31} &\textbf{0.23}&0.04 & -0.02 \\
        LAD &0.03 &-0.05 &  -0.07& 0.10 &0.05 &-0.13 &0.06 & 0.00 \\
         AxS &0.02 &-0.05 &  -0.01& 0.06 &-0.11 &0.02 &-0.06 & -0.15\\ 
         \bottomrule
     \end{tabular}
     }
     \caption{Global Performance Correlation between the shift importance assigned to the corrupted feature $j$ and the actual performance change caused by missing values. In all scenarios, XPE and XPPE consistently demonstrate superior performance in indicating the impact of the corruption most reliably via the allocated shift attribution.}
    \end{table}

\section{Discussion and Conclusion}
We introduced Explanatory Performance Estimation (XPE) as a novel and agnostic framework to attribute an anticipated change in model performance induced by a distribution shift to individual features. Our approach requires no ground truth labels during deployment, which corresponds to the typical situation in practice and is applicable to any monitoring situation. We also provided a theoretical analysis and extensive experiments covering nine datasets to demonstrate the empirical success of our method across ten different shifts and three data modalities.\\
Our approach is limited by the capabilities of optimal transport to model the true observed shift based on samples, which can be challenging for high-dimensional data and may require computing the transport based on lower-dimensional input representations (as exemplified in the audio experiment) or the use of more sophisticated cost functions \cite{pmlr-v202-cuturi23a}. Another way to enhance scalability for individual applications is to leverage
model or data-specific approximations of Shapley Values \cite{chen2023algorithms} or to incorporate additional concepts from unsupervised domain adaptation \cite{kouw2019review} for aligning source and target samples. Moreover, explaining the effect of a shift in terms of individual input shifts might not always be the optimal level of abstraction. Hence, combining XPE with concept-based explanations \cite{yeh2022human} would be a natural extension. This could even further increase the applicability of XPE as an effective method to enable explanatory model monitoring in practice.

\bibliographystyle{ACM-Reference-Format}
\balance
\bibliography{bibliography_final}

\appendix
\section{Mathematical Proofs}
In this section, we provide the proof of Theorem 5.2 in the main paper. For the definition of $\varepsilon$-approximate label-preserving distribution shifts, we refer to Definition 5.1 in Section 5. To start, we introduce a well-known equivalent characterization of the total variation distance as given for instance in \cite{gibbs2002choosing}.
\begin{lemma} Let $\mathcal{X}$ be a measurable space and $\mu$ and $\nu$ two probability measures on $\mathcal{X}$. Then 
\begin{align*} 
2\; d_{TV}(\mu,\nu)= \underset{h, \lVert h \rVert_{\infty} \le 1}{\max} \lvert \int_{\mathcal{X}} h d\mu - \int_{\mathcal{X}} h d\nu \rvert
\end{align*} where $h$ specifies any measurable function $h: \mathcal{X} \rightarrow \mathbb{R}$.
\end{lemma}
Based on this we can finally conduct the proof of Theorem 5.2.
\begin{proof}\textit{(Theorem 5.2)}  The true performance of a model $f$ in the target domain $\mathcal{L}_t$ and the estimated performance based on label transport $\widehat{\mathcal{L}_t^T}$ are given by:
\begin{align*}
    \mathcal{L}_t & = \mathbb{E}_{P_t(X_t,Y)}\left[\mathcal{L}(f(X_t), Y) \right] \\
    \widehat{\mathcal{L}_t^T} & = \mathbb{E}_{P_s(X_s,Y)}\left[\mathcal{L}(f(T(X_s)), Y) \right]  
\end{align*}
Then we have:
\begin{align*}
    &| \mathcal{L}_t-  \widehat{\mathcal{L}_t^T} | = \\
    &=\lvert\mathbb{E}_{P_t(X_t,Y)}\left[\mathcal{L}(f(X_t), Y) \right] -\mathbb{E}_{P_s(X_s,Y)}\left[\mathcal{L}(f(T(X_s)), Y) \right]\rvert \\
    &=|\mathbb{E}_{P_t(X_t)}\mathbb{E}_{P_t(Y|X_t)}\left[\mathcal{L}(f(X_t), Y) \right] - \\
    & \qquad\mathbb{E}_{P_s(X_s)}\mathbb{E}_{P_s(Y|X_s)}\left[\mathcal{L}(f(T(X_s)), Y) \right]| \\
    &\overset{(1)}{=}|\mathbb{E}_{P_t(T(x_s))}\mathbb{E}_{P_t(Y|T(X_s))}\left[\mathcal{L}(f(T(X_s)), Y) \right] - \\    &\qquad\mathbb{E}_{P_s(X_s)}\mathbb{E}_{P_s(Y|X_s)}\left[\mathcal{L}(f(T(X_s)), Y) \right]| \\
    &\overset{(2)}{=}|\mathbb{E}_{P_s(X_s)}\mathbb{E}_{P_t(Y|T(X_s))}\left[\mathcal{L}(f(T(X_s)), Y) \right] - \\    &\qquad\mathbb{E}_{P_s(X_s)}\mathbb{E}_{P_s(Y|X_s)}\left[\mathcal{L}(f(T(X_s)), Y) \right]| \\
    &\overset{(3)}{\le}\mathbb{E}_{P_s(X_s)}\big(|\mathbb{E}_{P_t(Y|T(X_s))}\left[\mathcal{L}(f(T(X_s)), Y) \right] - \\    &\qquad\qquad\mathbb{E}_{P_s(Y|X_s)}\left[\mathcal{L}(f(T(X_s)), Y) \right]| \big)  \\
    &\overset{(4)}{\le}\mathbb{E}_{P_s(X_s)}\Big[\; \underset{h, \lVert h \rVert_{\infty} \le 1}{\max}\Big\{|\mathbb{E}_{P_t(Y|T(X_s))}\left[h(X_s,Y) \right] - \\    &\qquad\qquad\mathbb{E}_{P_s(Y|X_s)}\left[h(X_s, Y) \right]|\Big\} \Big]  \\
    &\overset{(5)}{\le}\mathbb{E}_{P_s(X_s)}\big[2d_{TV}\big(P_t(Y|T(X_s)),P_s(Y|X_s)\big) \big] \overset{(6)}{\le} 2\varepsilon 
\end{align*}
\end{proof} In step (1) we performed a change of variables using $x_t = T(x_s)$ and in (2) we used property 1 of being $\varepsilon$-approximate label-preserving. In step (3) we used Jensen's inequality to pull the absolute value inside the expectation and in step (4) we used the assumption that $\lVert\mathcal{L} \lVert_{\infty} \le1$. In step (5) we applied the Lemma specified above and in step (6) we leveraged property 2 of being $\varepsilon$-approximate label-preserving.

\section{Details on Experiments}
\paragraph{Data and Models}
For the image data experiments, we analyzed MNIST \cite{lecun1998gradient}, FashionMNSIT \cite{xiao2017fashion} as well as OrganaMNIST and PneumoniaMNIST from the MedMNISTv2 benchmark \cite{medmnistv2}. For each dataset we trained and evaluated a LeNet model \cite{lecun1998gradient}. All models have been trained for 100 epochs with early stopping based on a patience of 10 epochs and PyTorch's Adam optimizer with a batch size of 16 and a learning rate of 1e-3. The used audio data set \cite{audiomnist2023} is publicly available. We downsampled the original raw sound data to a new sampling frequency of 2048 Hz and generated a spectrogram for every sound file using Short-time Fourier Transform (STFT). The STFT is calculated using a segment length of 455 and a overlap of 445. The training and test sets each consist of 12 randomly drawn participants according to the by the experiment determined gender distribution. According to the original paper \cite{audiomnist2023} of the audio data set we also used AlexNet\footnote{Used PyTorch implementation of AlexNet: https://github.com/Lornatang/AlexNet-PyTorch} \cite{AlexNet} to solve the digit classification task. The models are trained for 40 epochs. The training has been performed using an Adam optimizer with a batch size of 16 and a learning rate of 1e-3. For the tabular data, we selected four classification datasets and downloaded the preprocessed version provided by \cite{grinsztajn2022tree}. For each dataset we performed a 80-20 train-test split and fitted an XGBoost classifier with 50 estimators of max depth 5 as well as a Multi-Layer Perceptron (MLP) with one hidden layer of size 128.
\paragraph{Optimal Transport} In all experiments we match the same amount of samples in source and target domain using EMDTransport solver in the POT library \cite{flamary2021pot} with default parameters. Thus, the estimated couplings typically result in a one-to-one matching and the subsequent estimation of XPE and XPPE based on $\hat{T}$ are equivalent to those leveraging the entire coupling $\pi$.
\paragraph{Shift Attribution}
To compute the necessary Shapley Values for each shift attribution method we relied on the model-agnostic KernelSHAP implementation provided by \cite{lundberg2017unified}. All Shapley Values have been computed using a sample size of $3000$ and the default choices for all other hyperparameters. For LAD and AxS baselines we used a background dataset of $30$ random samples for tabular data and a fixed baseline of zeros otherwise. For the AxS metric, we estimated the shift mask $M^{\textit{KS}}$ using a featurewise two-sided Kolmogorov-Smirnov test with 95\% confidence threshold to signal a shift as provided by \cite{alibi-detect}.
\paragraph{Shift-Faithfulness and Complexity} To compute the metrics we relied on the implementation provided by \cite{hedstrom2023quantus}. For Shift-Faithfulness, we used the Faithfulness Correlation metric, specified the perturbation baseline to be the estimated per-shift version of each sample and chose a sample size of 100 with a subset size $|K|$ of 64. All other hyperparameters correspond to their default choices. Moreover, we removed samples where the anticipated performance change is only marginal as this causes either all shift attributions to be zero or causes numerical problems during the computation of the correlation value used in Shift-Faithfulness.
\paragraph{ROAR-S Metric}
Our implementation of Remove, Retrain, and
Shift ROAR-S is based on the implementation by \cite{khakzar2021neural} of the ROAR metric \cite{hooker2019benchmark}.
Given the high computational requirements of ROAR, we sub-sample the original datasets to obtain  $ D_s^{\textit{train}}$ and $D_s^{\textit{test}}$ with $N_{ROAR}=1000$ samples each. $D_s^{\textit{train}}$ is used to train the pre-removal LeNet model $f$. For each shift we create the shifted dataset versions $D_t^{\textit{train}}$ and $D_t^{\textit{test}}$ and compute the different feature shift attributions $\phi^{\textit{Shift}}$. Next we remove the top 5\% of the features attributed the highest importance. This yields the post-removal sets $\tilde D_s^{\textit{train}}$, $\tilde D_s^{\textit{test}}$, $\tilde D_t^{\textit{train}}$ and $\tilde D_t^{\textit{test}}$. The new source training set $\tilde D_s^{\textit{train}}$ is used to retrain the model yielding $\tilde f$. With cross-entropy loss $\mathcal{L}$, the test performances $\mathcal{L}_t^{\textit{test}} = \frac{1}{N_{\textit{ROAR}}}\sum_{(x,y)\in D_t^{\textit{test}}} \mathcal{L}(f(x), y)$, $\mathcal{L}_s^{\textit{test}} = \frac{1}{N_{\textit{ROAR}}}\sum_{(x,y)\in D_s^{\textit{test}}} \mathcal{L}(f(x), y)$ of the model $f$ without removal and the test performances $\tilde {\mathcal{L}}_t^{\textit{test}} = \frac{1}{N_{\textit{ROAR}}}\sum_{(x,y)\in \tilde D_t^{\textit{test}}} \mathcal{L}(\tilde f(x), y)$, $\tilde{\mathcal{L}}_s^{\textit{test}} = \frac{1}{N_{\textit{ROAR}}}\sum_{(x,y)\in \tilde D_s^{\textit{test}}} \mathcal{L}(\tilde f(x), y)$ after removal on the retrained model $\tilde f$, the ROAR-S score can be defined as:
\begin{align*}
    \text{ROAR-S}(\tilde{\mathcal{L}}_t^{\textit{test}},\tilde{\mathcal{L}}_s^{\textit{test}},\mathcal{L}_t^{\textit{test}},\mathcal{L}_s^{\textit{test}})=
    \frac{\max(0,\tilde{\mathcal{L}}_t^{\textit{test}}- \tilde{\mathcal{L}}_s^{\textit{test}})}{\mathcal{L}_t^{\textit{test}} -  \mathcal{L}_s^{\textit{test}}} 
\end{align*}
All models have been trained for 100 epochs with early stopping based on a patience of 10 epochs, an Adam optimizer with a batch size of 16 and a learning rate of 1e-2.

\end{document}